 \documentclass[sigconf,authorversion,nonacm]{acmart}
\usepackage{array}

\usepackage{graphicx} 
\usepackage{subcaption} %
\AtBeginDocument{%
  \providecommand\BibTeX{{%
    \normalfont B\kern-0.5em{\scshape i\kern-0.25em b}\kern-0.8em\TeX}}}

\setcopyright{acmcopyright}
\copyrightyear{2018}
\acmYear{2020}
\acmDOI{10.1145/1122445.1122456}

\acmConference[Malta '20]{Malta '20: 11th Workshop On Procedural Content Generation (PCG 2020) }{September 15--18, 2020}{Malta}
\acmBooktitle{Malta '20: 11th Workshop On Procedural Content Generation (PCG 2020), September 15--18, 2020, Bugibba, Malta}

\begin{document}

\title{Procedural Content Generation of Puzzle Games using Conditional Generative Adversarial Networks}


\author{Andreas Hald}
\affiliation{%
  \institution{IT University of Copenhagen}
  \city{Copenhagen}
  \country{Denmark}}
\email{andreas_hald@protonmail.com}

\author{Jens Struckmann Hansen}
\affiliation{%
  \institution{IT University of Copenhagen}
  \city{Copenhagen}
  \country{Denmark}}
\email{jsh_mailbox@protonmail.com}

\author{Jeppe Kristensen}
\affiliation{%
  \institution{IT University of Copenhagen}
  \city{Copenhagen}
  \country{Denmark}}
\email{jeppek@tactile.dk}

\author{Paolo Burelli}
\affiliation{%
  \institution{IT University of Copenhagen}
  \city{Copenhagen}
  \country{Denmark}}
\email{pabu@itu.dk}




\begin{abstract}
    In this article, we present an experimental approach to using parameterized Generative Adversarial Networks (GANs) to produce levels for the puzzle game Lily’s Garden\footnote{https://tactilegames.com/lilys-garden/}.
    We extract two condition-vectors from the real levels in an effort to control the details of the GAN's outputs. \\
    While the GANs performs well in approximating the first condition (map-shape), they struggle to approximate the second condition (piece distribution). We hypothesize that this might be improved by trying out alternative architectures for both the Generator and Discriminator of the GANs.
    
    
\end{abstract}



\maketitle


\section{Introduction}
    PCG can  be  defined  as  the  generation  of  content  for  a  game  by  algorithms  which  require  little  to  no  supervision  by  a  game-designer \cite[p. 1]{Shaker2016}. As content production is an often tedious and time-consuming task, game-companies are increasingly utilizing PCG in an effort to alleviate some of the pressure of having human designers create all of the content within games. But content production often relies on domain specific knowledge - in order to create levels that are playable and continually enjoyable as they enter into the customers overall experience of the game - and in light of this requirement it is often challenging to automate the entirety of the content production with algorithms. However, this is exactly what researchers are attempting in a relatively new paradigm in PCG called Procedural Content Generation via Machine Learning (PCGML).  
    
    In recent years we have seen studies in PCGML using Generative Adversarial Networks (GAN). However,
    GANs are generally not great at upholding the constraints needed for necessary content used in games \cite{Torrado2019}.
    Previous studies have used different methods to overcome the challenges of keeping the content functional for games by using evolutionary algorithms and capturing information about spatial relationships in levels \cite{Volz2018, Torrado2019}.  
    In this study we are drawing inspiration from mixed-initiative generation where humans and computers interact to pull the content in a certain direction, specifically computer-aided design tools like Sentient Sketchbook \cite{Shaker2016, Liapis2013}.

    \indent A natural extension to our work could therefore be to create a tool that can aid the content producers in speeding up the process of generating game-levels for the puzzle-game Lily's Garden. To do this we present a preliminary study of using Conditional Generative Adversarial Networks to generate a draft of levels based on the variables of the game that we believe to be domain specific.



\section{Related Work}

\subsection{Procedural Content Generation}
Procedural Content Generation (PCG) focuses on using algorithms to produce various types of content that is used for games \cite[p. 1]{Shaker2016}. 
Content include levels, quests, textures and the PCG methods used are usually specialized in doing one particular type of task, but Shaker et. al have also suggested multi-content PCG, as an alternative path \cite[p. 5]{Shaker2016}. 

Throughout the existence of PCG the motivation behind using it has changed. In the early days of video games PCs didn't have a lot of storage space, so instead of saving different game layouts, they could be randomly generated \cite[p. 4]{Shaker2016}. Today, game development for AAA games has become more expansive which requires more designers and time. As such, if algorithms alleviate the need for increasing staff or help make designers more efficient it would certainly be beneficial, which is also true for smaller teams \cite[p. 3]{Shaker2016}. Another aspect of today's use of PCG is the possibility of personalizing content for players to keep them engaged, which could span from different scenarios that a player enjoys, to the difficulty or patterns in a platform game that a player keeps going back to \cite{Togelius2007, Shaker2009}. 

\indent In recent years a new paradigm in PCG has emerged, Procedural Content Generation via Machine Learning (PCGML) \cite{Summerville2017}. This paradigm separates itself from PCG in the sense that while PCG approaches use machine learning models, PCGML samples directly from the model, based on existing distributions of content \cite{Summerville2017}. Examples of this paradigm includes Snodgrass \& Ontañón's work, in which they use Constrained Multi-Dimensional Markov Chains in order to control the generation of both levels for the games Super Mario Bros and Kid Icarus \cite{Snodgrass2015}. Another example is Sarkar et. al's. use of Variational Autoencoders for Controllable Level Blending between Games \cite{Sarkar2017}, Sarkars work is especially interesting in that they seek to optimize for multiple different features, which has a lot of commonality with the line of inquiry that we are pursuing.

This consequently is the paradigm of PCG that we are working under, with a focus on Generative Adversarial Networks which we will briefly describe next.

\subsection{Generative Adversarial Networks}
Generative Adversarial Networks (GAN) is a collection of architectures introduced by Ian Goodfellow et al. \cite{Goodfellow2014}. in 2014. The basic idea is that two neural networks, a generator and discriminator, are competing against each other in a mini-max game, where the generator tries to fool the discriminator into predicting its output as real, and the job of the discriminator is to distinguish the fake samples of the generator from real samples of data \cite{Goodfellow2014, Goodfellow2016}. The two models are trained simultaneously and theoretically the mini-max game should lead to convergence. However, there are multiple challenges when balancing the architecture which has been a focus point in the continued research on the subject \cite{Goodfellow2016}. Multiple improvements has been made to the GAN architecture, such as the Deep Convolutional GANs \cite{Radford2015}, which is broadly considered a general improvement on the original model.

\subsection{PCGML and GAN}
Since the GAN architecture was presented in 2014, we are aware of three articles using GANs to generate content for games  \cite{Giacomello2018, Volz2018, Torrado2019}. 
Giacomello et al. (2018) use levels from the original DOOM by extracting topological features and level images to generate new levels with two different GANs-architectures. The purpose is to evaluate the networks performance in generating levels similar to human designed ones, with both training on level images and one conditioned on the topological features as well \cite{Giacomello2018}. 

\indent Volz et al. (2018) train a GAN on mario levels and uses an evolutionary algorithm on the latent space to improve properties relating to difficulty, and playability is also checked using a player agent \cite{Volz2018}.

\indent Torrado et al. (2019) aim to mitigate common problems in PCG like data scarcity and generating functional content for games by implementing Self-attention to make sure that objects required for functionality (in their case a key and door) is to be found within a synthetic level, and adding playable examples of generated content to the training loop for their GAN \cite{Torrado2019}.

\indent Our approach uses some of the same ideas shown in the aforementioned articles, but we explore them in the context of possibly creating a mixed-initiative design-tool between GANs and game level designers at a later stage.



\section{Method}
    
    In this section, we will first give the reader an impression of what the game \textit{Lily's Garden} is about. Secondly we will describe the way that we represent the game levels as data to be used in GANs. Thirdly we discuss the important variables, specific to the puzzle-game genre, and finally we describe the GAN-algorithms that we have implemented in our project.   
    
    \subsection{A Lily's Garden tutorial}
    
        \begin{figure}
            \centering
            \includegraphics[width=8.2cm, height=4.2cm]{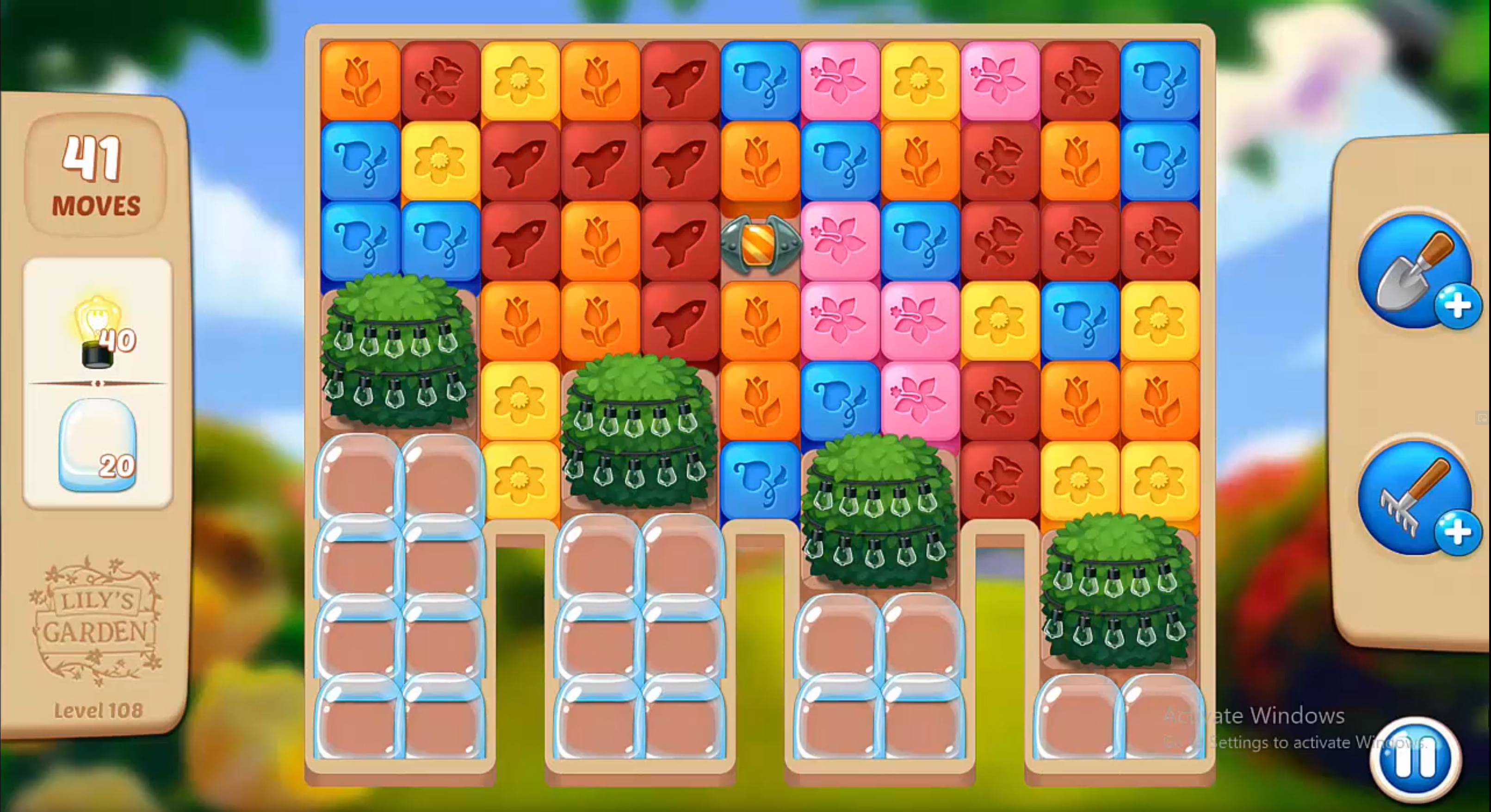}
            \caption{Lily's Garden level 108 at the initial state.}
            \label{fig:teaser}
        \end{figure}

        \begin{table}[ht]
            \caption{Table viewing a Term, its visualization, and whether or not it is clickable.}\vspace*{-6pt}
            \begin{tabular}{|c|c|c|}
            \hline
            \textbf{Unique Pieces} & \includegraphics[scale=0.17]{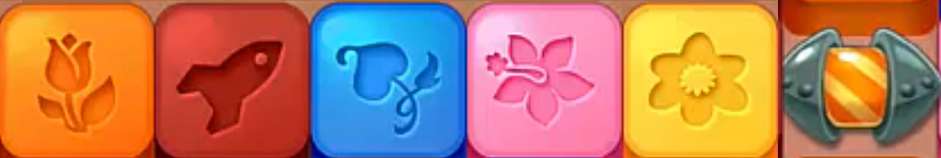}\includegraphics[scale=0.16]{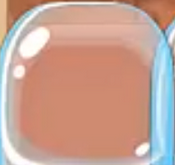}\includegraphics[scale=0.11]{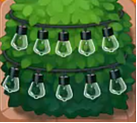} & \textit{Clickable}\\
            \hline
            \textbf{Color Island} & \includegraphics[scale=0.25]{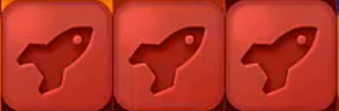}&\textit{Clickable}\\
            \hline
            \textbf{Blockers} & \includegraphics[scale=0.18]{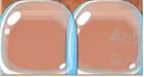}\includegraphics[scale=0.11]{blocker2.png}&\textit{Non-Clickable}\\
            \hline
            \end{tabular}
            \label{tab:gt}
        \end{table}


    \begin{figure*}[h]
                \centering
                \includegraphics[width=10cm, height=4cm]{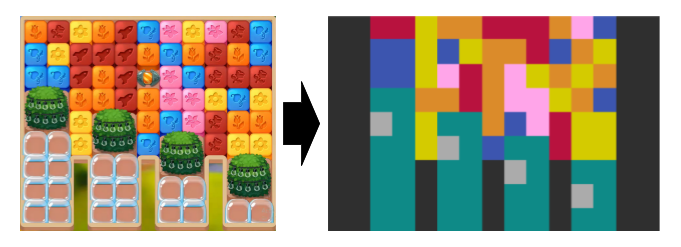}\vspace*{-10pt}
                \caption{Reduced Representation of level 108}
                \label{reducedlvl}\vspace*{-12pt}
            \end{figure*}

  \begin{figure*}
                \includegraphics[width=12cm, height=2cm]{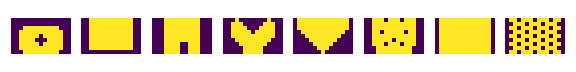}\vspace*{-15pt}
                \caption{8 randomly selected level-shapes from Lily's Garden, visualized as heatmaps.}
                \label{fig_level_shapes}
            \end{figure*}

        Lily's Garden is a puzzle game where the player has to collect a certain amount of the \textit{pieces} located on the game-board (see left column in fig. \ref{fig:teaser}). Collecting pieces can be done by connecting two or more \textit{clickable}-pieces of the same color (For example the 7 red connected pieces in the top left  quadrant of fig. \ref{fig:teaser}) which makes them clickable (referred to as \textit{color-islands}). However, some pieces are not themselves clickable (referred to as \textit{blockers}).
        This is for example the bushes in fig. \ref{fig:teaser} and in this case, the only way to collect them is by clicking on a color-island touching upon the non-clickable piece. While there are some other categories of pieces, the color-islands and blockers are the most important types and we will limit ourselves to focusing on those for the remainder of the article.

    \subsection{Data representation}
        
        \subsubsection{The Representation Challenge:} 
            Within PCGML there isn't a general agreement on how to represent the game structure \cite[p.11]{Summerville2017}, but most examples we have seen seem to represent each distinct piece in a game-level by a distinct channel within a 3D matrix \cite{Giacomello2018, Torrado2019, Volz2018}. However, Lily's Garden has 56 unique pieces and vastly more combinations of unique pieces stacked upon each other. By comparison (and to our understanding) Volz et. al. \cite{Volz2018} represented an entire Super Mario Bros level with just 10 unique piece representations. 
        
        \subsubsection{Reducing the number of channels:}
            In order to reduce the complexity of cleaning, visualizing and analysing the data, we therefore decided to reduce the data representation into 8 distinct piece-types: (1) Cell-layer, indicating the levels shape, and where other unique pieces are allowed to be placed. (2) Blocker-layer (3)-(8) color-layers 1-6. - 
            We realize doing this is to reduce the problem we were originally trying to solve, but an important feature of the human-made levels are, that the main way to \textit{start} a level is by clicking on a color-island and thus the basic but important feature of \textit{startability} is maintained.

Fig. \ref{reducedlvl} shows the reduced representation which has some key differences from the original. The bubble pieces are not visible in our representation and the \textit{bushes} are represented with 1 blocker piece, you might also notice that the piece placement is not exactly the same, this is because the game state are generated on different random seeds.
            The first two dimensions of the levels are originally (9,13), but we increase it to (9,15) by padding with empty values. This is done to be able to apply transposed convolutional layers in our generator, which makes it possible to go from (3,5) up to the (9,15) size. The third dimension is the number of unique pieces in our representation, which gives us a final multi-channel representation of 15x9x8.

    \subsection{Important variables}\label{important_variables}
    
        Every game has some defining variables which are important to the functionality of a level and further the player experience. In this section we will briefly mention those that we consider important in Lily's Garden. 
        
        \subsubsection{Shape:} 
            Shape is important because it determines where pieces can spawn on the screen and further because it limits where the player can interact with the level.

            It is for this reason that an entire unique piece-type in our data-representation is indicatory of the level shape (the cell-layer). Additionally, given the many varieties of shapes there exists within the human made levels (see fig. \ref{fig_level_shapes}) we found it reasonable to assume that in a PCG-context, a variable that game-designers might want to control is the level-shape.
            
        \subsubsection{Piece Distribution} 
\looseness-1            The distribution of pieces in a level is also an important variable. In fig. \ref{fig:teaser} for example, it is evident that some pieces, such as the red- and orange-cookies (henceforth referred to as clickable-types), are more dominating than the pink or the blockers underneath. In Lily's Garden there generally are a lot of unique pieces and while the most re-occurring pieces are the clickable-type, a level is often dominated by clickable- and some \textit{particular} blocker types. While the clickable pieces are typically clustered into color-islands - with some additionally lonely pieces being scattered across the board - the blockers typically make up some local shapes within the global shape. Given their visual importance, as well as role in the functionality of a level, we also found the piece-distribution a plausible variable that game-designers might want to be able to control.
    
    \subsection{Train and test set}
        We had access to 655 unique levels of Lily's Garden. But to increase the amount of data even further, we flipped the levels horizontally, vertically and diagonally in each channel, giving us 2620 levels. We then used 85\% of the levels for training and 7.5\% (196 levels) for test and validation set respectively. While we do not use the validation set in this article, they are stored for possible future work. 
    
    \subsection{GAN architectures}
        
        \begin{figure*}
                \includegraphics[width=14cm, height=7cm]{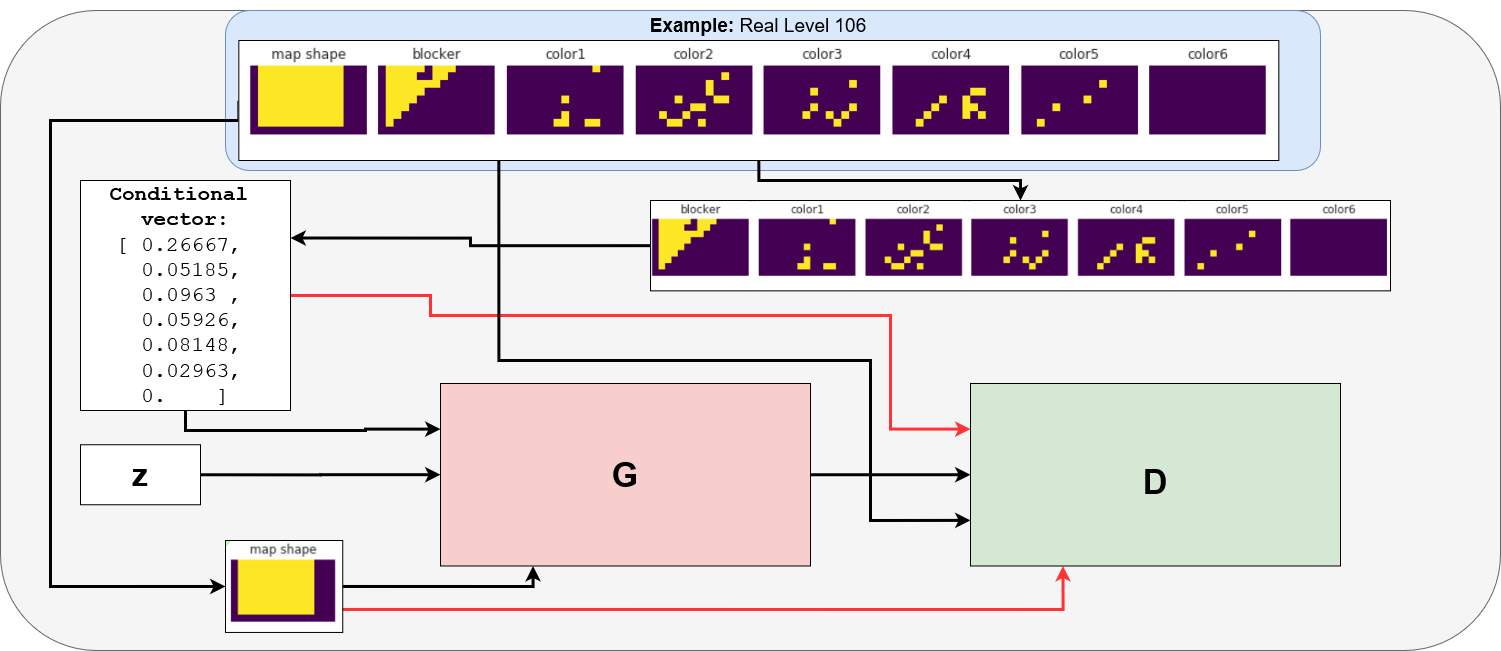}
                \caption{WGAN-GP-PE and CWGAN-GP, the red lines are only included in the CWGAN-GP model. The level-shape and the piece-distribution of each level is extracted as conditions. The piece-distribution-vector is generated by counting the amount of 1's in the blocker- and color-channels and assigning each value to an index in the condition vector. Each number is then divided by 135 (total possible size of a level) to normalize the values between 0-1, indicating the proportion of the level that a particular channel inhabits.}
                \label{fig_model_architecture}
        \end{figure*}
    
        In this article, we have developed two types of GANs: 1. The \textit{Wasserstein GAN using Gradient Penalty with Parametric Embeddings} (WGAN-GP-PE) and the \textit{Conditional Wasserstein GAN using Gradient Penalty} (CWGAN-GP). The exact architecture of the generators and critics can be accessed here\footnote{https://github.com/DresRumler/pcg-workshop-visualizations}. Let us unpack these models a few terms at a time: 
        
        \subsubsection{Wasserstein and Gradient Penalty:} 
            As the name may suggest we utilize the wasserstein-loss function which was suggested by Arjovsky et. al. as a general improvement over cross-entropy and leads to more stable training of GANs \cite{Arjovsky2017}. Further we use \textit{Gradient Penalty} as suggested by Gulrajani et. al. as a general improvement to the wasserstein-loss, stabilizing the gradients and theoretically approximating the real data distribution better \cite{Gulrajani2017}. In practice, this also proved to be the case for us. 
        \subsubsection{Conditonal and Parametric Embeddings:}
            Traditionally GANs \cite{Goodfellow2014} only get the random-vector \textbf{z} (drawn from some normal distribution) as an input-vector. However, if one wants to control the output of the generator towards producing specific types of images, one has to introduce conditions as suggested by Mirza et. al. \cite{Mirza2014}. In our case, the conditions are the level-shape and piece-distribution, which we described in section \ref{important_variables}. As can be seen 
            in fig. \ref{fig_model_architecture} we use both variables as input to the model. In the case of the WGAN-GP-PE we only feed the generator the variables (the red-lines in fig. \ref{fig_model_architecture} are not included), where in the CWGAN-GP we also feed them to the discriminator (the red-lines in fig. \ref{fig_model_architecture} are included). This is the only difference between the models, but given that the discriminator is actually not taking the conditional vectors into consideration in the WGAN-GP-PE-model, we decided to define the conditional vectors in this case as \textit{Parametric Embeddings}.
            
        \subsubsection{Hyperparameters}
            \begin{itemize}
                \item Loss-function: Wasserstein Loss.
                \item Critic trained 5 times for every 1 time Generator is trained.
                \item Optimizer: Adam with learning rate 0.0001, beta term 1 and 2 equal to 0.5 and 0.9 respectively.
                \item Batch-size: 32 levels.
                \item Epochs: 300
            \end{itemize}
        



\section{Experiments} \label{sect_4}
    
    Most GAN-literature has concerned itself with producing realistic looking images of human-faces\footnote{http://mmlab.ie.cuhk.edu.hk/projects/CelebA.html} and more generally objects from the ImageNet\footnote{http://image-net.org/}. As such, the most used evaluation metrics, such as the Inception-score \cite{Salimans2016} and the Fréchet Inception Score \cite{Heusel2017} is also mainly useful in this context of measuring how realistic looking an image is.

To judge the quality of a game-level however - and specifically for Lily's Garden - is quite a different task and we have therefore run 7 distinct experiments which seeks to test the quality of our generators from numerous different angles. 
    
    \subsection{Data distribution approximation of the generators}
    
        Firstly, we investigate how well our generators has approximated the data distribution of the training set, by looking at the proportion of pieces that the generated data sets and training set contain respectively. We have generated 250 samples for each level in the test set (of 196 levels) Which gives us a total of 49.000 synthetic levels for each of the generators.
         
        \begin{figure*}
            \includegraphics[width=14cm, height=5cm]  {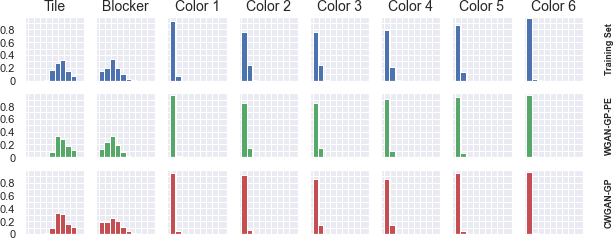} 
            \caption{Distributions of the Training Set, WGAN-GP-PE and CWGAN-GP respectively. X-axis ranges between 0 and 135 and Y-axis is the proportion of piece-amount that falls within a given range.}
            \label{fig_hist_distributions}
        \end{figure*}
        
        From fig. \ref{fig_hist_distributions} and tab. \ref{tab_median_and_standard_error}, it is clear that both generators have approximated the training distribution pretty well with regards to the number of pieces that each cell-layer contains.
        However, the color layers seem to be on the low side of the training-set consistently, while the cell-layer is slightly over represented. 
       
        \begin{table}[H]
          \caption{0.5'th quantile and standard error}
            \label{tab_median_and_standard_error}
            \begin{tabular}{|c|c|c|c|}
                \toprule
                Layer & Training Set & WGAN-GP-PE & CWGAN-GP \\ \midrule
                Cell & $81\pm_{14}^{17}$  & $84\pm_{14}^{18}$ & $84\pm_{15}^{17}$ \\ [2pt]
                Blocker & $34\pm_{19}^{17}$ & $32\pm_{17}^{16}$ & $33\pm_{22}^{22}$  \\ [2pt]
                Color & $0\pm_{0}^{11}$  & $0\pm_{0}^{4}$ & $0\pm_{0}^{9}$   \\ [2pt]
                Color 2 & $9\pm_{9}^{6}$  &$ 6\pm_{4}^{7}$ & $5\pm_{3}^{6}$  \\ [2pt]
                Color 3 & $9\pm_{5}^{6}$  &$ 7\pm_{4}^{6}$ & $7\pm_{4}^{6}$  \\ [2pt]
                Color 4 & $ 9\pm_{5}^{5}$  & $6\pm_{3}^{5}$ & $7\pm_{4}^{6}$  \\ [2pt]
                Color 5 & $6\pm_{6}^{6}$ & $4\pm_{3}^{5}$ & $4\pm_{3}^{5}$   \\ [2pt]
                Color 6 & $0\pm_{0}^{5}$ & $1\pm_{1}^{4}$ & $0\pm_{0}^{3}$   \\ [2pt]
                \bottomrule
            \end{tabular}
        \end{table}
        
        The generators must be said to have come close to the training-data distribution with respect to the proportion of pieces, but whether this translates into levels that share commonalities to the training set is what we will turn our attention to now.
    
    \subsection{Visual Quality}
        
        In this experiment, we compare the training set levels, to the closest and farthest synthetic levels measured in wasserstein-distance, the same metric used for the generators cost-function.
        
        \begin{figure*}
            \centering
\begin{subfigure}[h]{0.47\textwidth}
               \includegraphics[width=\textwidth]{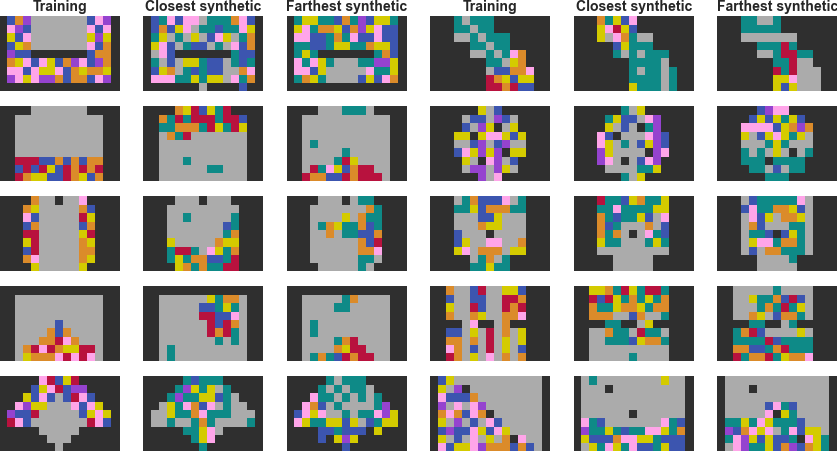}\caption{WGAN-GP-PE}\label{fig_visual_levels_wgan}
\end{subfigure}
\hfill
\begin{subfigure}[h]{0.47\textwidth}
               \includegraphics[width=\textwidth]{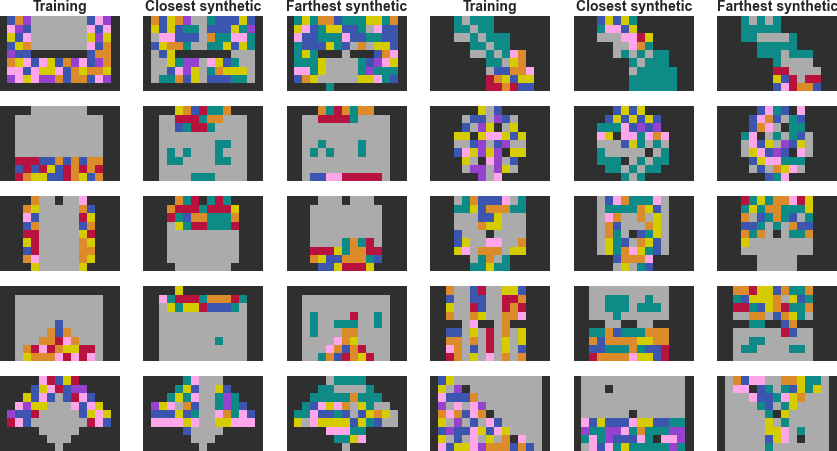}\caption{CWGAN}\label{fig_visual_levels_cwgan}
\end{subfigure}%
            \caption{Test levels compared to the closest- and farthest synthetic level of the Generators in Wasserstein distance.}
             \label{fig:my_label}
        \end{figure*}
        
    

        \subsubsection{Piece distributions}
            The most obvious difference between the training- and synthetic-set examples in fig. \ref{fig_visual_levels_wgan} \& \ref{fig_visual_levels_cwgan}, is how representative \textit{empty} board pieces (represented with the green color) is. In the training-set examples, only 2 out of 10 levels contains empty pieces, whereas every synthetic-level contains some empty pieces. Considering that both generators has been trained on the piece-distribution vectors of the training-set, the presence of empty pieces in every synthetic level, suggests that the generators has failed to utilize the piece-distribution vectors as we intended. We will return to this observation in section \ref{sec_effect_of_piece-distribution-conditional vector}.
        
        \subsubsection{Symmetry}
            Another important aspect of the training levels, seems to be the symmetric smaller shapes within the level-shapes.
            Once again, both the generators seems to underperform in comparison to the training-set. However, the CWGAN-GP does seem to be a general improvement over the WGAN-GP-PE in this respect. A statement we base on the observations that: the synthetic levels of the CWGAN-GP has more clear-cut vertical and horizontal borders between the blocker-and color pieces. Further, the CWGAN-GP produces these pyramidic-shapes (see row 4, column3 (refered to as (4,3), and (5,6) in fig. \ref{fig_visual_levels_cwgan}, which seems to suggest that the model has also adapted well to diagonal lines (see fig.  \ref{fig_visual_levels_cwgan}).

    \subsection{Effect of level-shape-conditional vector - Likely Playable}
        In this experiment, we test how well our models are reproducing the level-shapes it is given as inputs. To do this, we extract the cell-layer of the produced levels and compare them to the original input level-shape. 
        
        \begin{figure}
                \includegraphics[width=196pt, height=36pt]{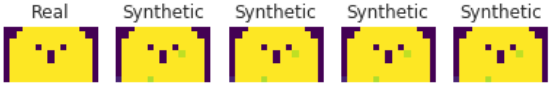}
                \caption{A test-set \textit{real} level-shape compared to the level-shape of 4 versions of the same synthetic levels of our CWGAN-GP}
                \label{fig_gen_level_shapes}
        \end{figure}
        
        From fig. \ref{fig_gen_level_shapes} it is clear that our CWGAN-GP approximates the real level shape quite well, albeit not perfectly. Notice that we have not rounded the synthetic values, and thus the green pieces in the images indicate where our CWGAN-GP is uncertain of whether to place a piece. While this mean our models are not perfect, it is worth noting how many of the pieces that the network is absolutely certain (1 on a scale of -1 to 1) should be placed in the spot.
        But how far from perfect are our models generally? To test this we generate 250 levels for each level-shape and condition-vector. We then count the number of times a level has failed and succeeded to fill in a desired spot (underfilled/overfilled) and report the average. The results are summarized in table \ref{tab_level_shapes}.
    
        \begin{table}[h]
          \caption{Average under\- and overfilling of the generators level\-shapes.}
            \label{tab_level_shapes}
            \begin{tabular}{ccc}
                \toprule
                 & WGAN-GP-PE & CWGAN-GP \\
                \midrule
                Avg. underfilled & 1.67 & 0.18 \\
                Avg. overfilled & 2.07 & 0.085  \\
                \bottomrule
            \end{tabular}
        \end{table}
        
        Table \ref{tab_level_shapes} shows us that the CWGAN-GP approximates the level-shape-condition much better than the WGAN-GP-PE, which indicates that feeding the level-shape conditional vector to the discriminator as well, has helped the generator in better approximating the real shape. Nevertheless, 18\% of the CWGAN-GP levels are underfilled and 8.5\% is overfilled and as such there is room for improvement.  
        
        
    \subsection{Effect of piece-distribution-conditional vector}\label{sec_effect_of_piece-distribution-conditional vector}
        In this experiment we test how well our models utilize the piece-distribution-vector. Recall that we feed a conditional-vector of size 7 (blocker- and color distributions) to the generator in both models and also the discriminator in our CWGAN-GP model. To test how it adjusts to the information in the vector, we will produce 100 new levels for each level in the test-set. We will then retrieve the distribution-vector for each of the produced levels and subtract the real distribution vector from the derived distributions-vectors. Ideally, the error should be close to 0, because this would mean that our models are recreating the distributions perfectly.
        
        \begin{figure*}
                \includegraphics[width=\linewidth]{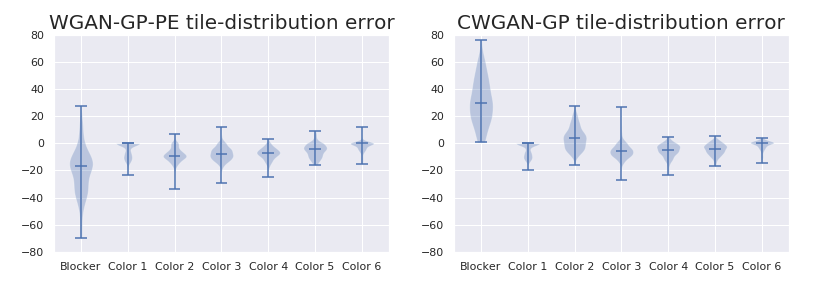}
                \caption{Violinplots of each index in the distribution-vector, showing the error between real- and vectors derived from generated levels. }
                \label{fig_tile_dist_comp}
        \end{figure*}
        
        
        From the violinplots in fig. \ref{fig_tile_dist_comp}  it does not seem like the models have approximated the conditional vector. However, we hypothesize that this might have to do with the fact that we reduced a lot of unique pieces into the single piece \textit{blocker} and thus our models might have learned that there are generally a lot of blocker-pieces, which it tries to distribute on every generated level. 
        
    \subsection{Testing Color-Islands}
    
        We test \textit{Color-islands} because it is indicative of whether there is anything clickable on a level to begin with and thus whether the level is \textit{startable}. We test this by producing 250 generated levels for each level in the test-set (a batch), and we then count the number of color islands within each level.
        
        \begin{figure*}
            \includegraphics[width=12cm, height=5cm]{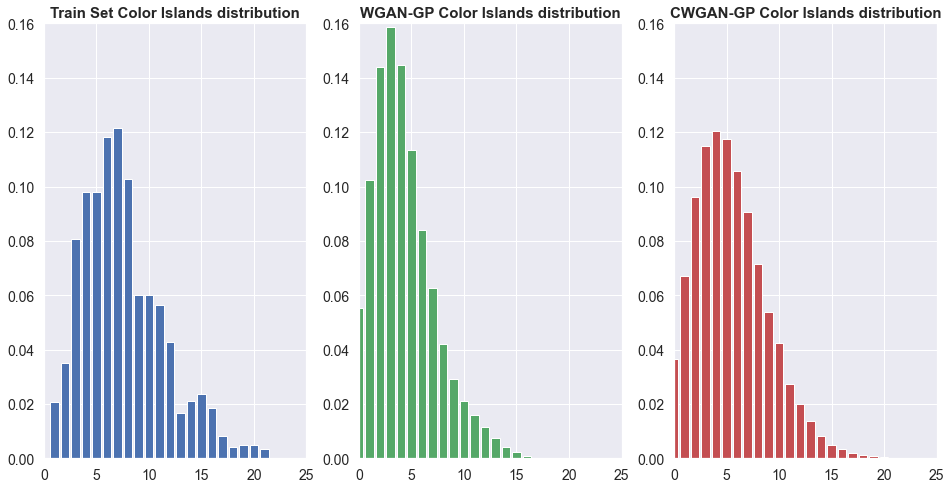}
                \caption{Bar Graphs showing the distribution of color-islands in train-set, WGAN-GP-PE- and CWGAN-GP synthetic-set respectively.}
                \label{fig_color_islands}
        \end{figure*}
    
        Fig. \ref{fig_color_islands}, shows that both the WGAN-GP-PE and CWGAN-GP has approximated the distribution of training-set closesly, although with some important differences. 
        \subsubsection{0 color-islands} From fig. \ref{fig_color_islands} it is clear that the minimum amount of color islands in a training-set level is 1. By contrast, the synthetic levels of the WGAN-GP-PE contains ca. 5.5\% levels with no color islands, and the CWGAN-GP contains a little less than 4\%. This effectively means that the generators both generate levels that are \textit{unstartable} and thus by extension \textit{unplayable}.
        
        \subsubsection{Largest proportions of color-islands} Both the generators tends to produce less color-islands than are in the training set. For the WGAN-GP-PE ca. 16\% of the levels has 3 color islands in them, but for the CWGAN-GP the highest proportion of levels are 12\% at 4 color islands. By contrast, the largest proportion (ca. 12\%) of levels in the training set has 7 color-islands in them.  
         
        
    \subsection{Testing Broken pieces}
        
        We test \textit{Broken pieces} by seeing if our model places any blocker- or color-piece outside the perimeter of the generated cell-layer. Like the preceding tests, this is done by producing batches of 250 generated levels and seeing how many of the levels within a batch are \textit{broken}. 
        
        \begin{figure*}
                \includegraphics[width=12cm, height=5cm]{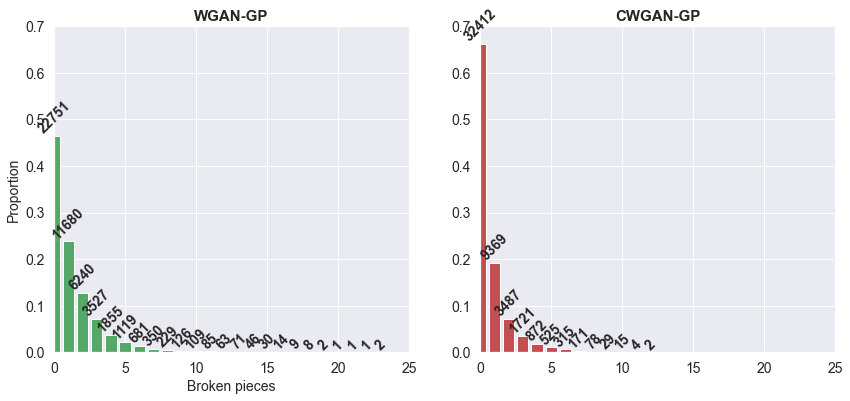}
                \caption{Bargraphs showing the proportion of n-broken pieces in the 49000 synthetic levels generated for WGAN-GP-PE and CWGAN-GP respectively.}
                \label{fig_broken_tiles}
        \end{figure*}
        
         Our WGAN-GP-PE produces 46\% levels that are not broken from the beginning, while our CWGAN-GP produces 66\% (see fig. \ref{fig_broken_tiles}). Evidently our generators struggles more with this test than the color-island test. 

 Interestingly, while we found that the piece-distribution vector did not contribute much to the actual distribution of tiles in the synthetic levels (see sec. \ref{sec_effect_of_piece-distribution-conditional vector}), the piece-distribution vector does seem to help improve the CWGAN-GP to generating 66\% non-broken levels from the beginning. Recall that the main difference between the WGAN-GP-PE and the CWGAN-GP, is that the discriminator network is also fed the piece-distribution vector in the CWGAN-GP. Thus while the piece-distribution vector is not necessarily working the way we intended, it does seem to effect the quality of the synthetic levels, when the discriminator also gets to take it into account.
 
    \subsection{Expressive Range}
    
        In this section, we test the expressive range \cite{Summerville2018} of the generators, in comparison to training set in two different experiments.
    
        \begin{figure*}
\includegraphics[width=12cm, height=5cm]{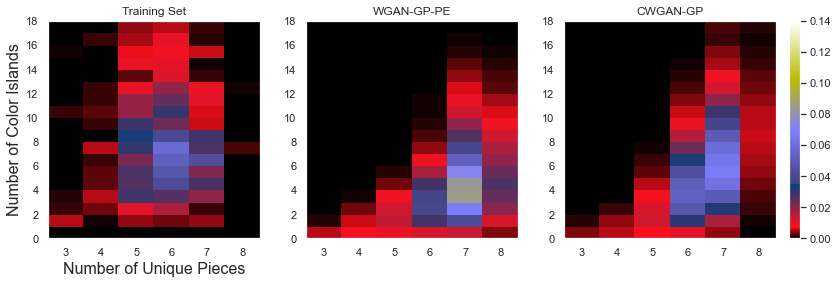}\vspace*{-6pt}
                \caption{2D Heatmaps, showing the expressive ranges of number of unique pieces and number color islands within the train- WGAN-GP-PE- and CWGAN-GP-set of levels respectively.}
                \label{fig_expressive_range_islands_unqiue}
        \end{figure*}
        
\begin{figure*}
 \includegraphics[width=12cm, height=5cm]{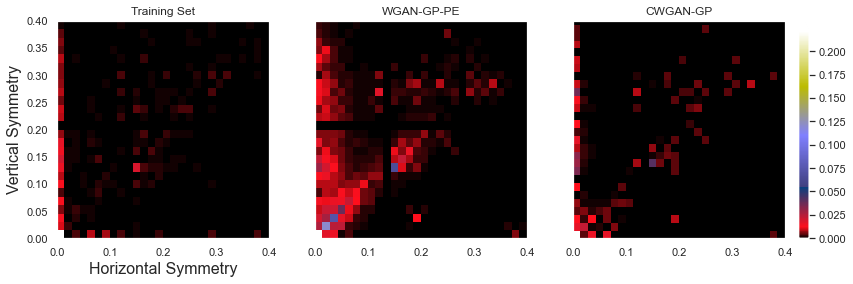}\vspace*{-6pt}
                \caption{2D Heatmaps, showing the expressive ranges of the horizontal- and vertical symmetry within the train- WGAN-GP-PE and CWGAN-GP-set of levels respectively.}
                \label{fig_expressive_range_symmetry}
        \end{figure*}
        
        Firstly, we test the generators expressive ability over the number of color islands and unique pieces in a level. Evidently, from fig. \ref{fig_expressive_range_islands_unqiue}, both generators tend to overestimate the number of unique pieces in a level, but the WGAN-GP-PE more so than the CWGAN-GP.
        The training-set is generally much more spread across the middle of the heatmap, than the generators, which tend towards the lower right corner with all colors represented but fewer color islands. In other words, the training set usually has 5-7 unique pieces, whereas both generators typically has 6-8, although the CWGAN-GP once again is closer to the training-set than the WGAN-GP-PE. 
        \enlargethispage{13pt}
        
        Secondly, we test the generators ability to create symmetrical levels by flipping the level-shapes (see fig. \ref{fig_expressive_range_symmetry}), first in the middle horizontal axis (X-coordinate) and secondly on the vertical (Y-coordinate). We then measure the hamming-distance \cite{data_mining} and use this as an indicator of how symmetrical a level is. Thus, if a level is both symmetrical on the X- and Y-axis, the distance is zero. Interestingly, the CWGAN-GP seems to adapt much better to the training-set, than do the WGAN-GP-PE. This is especially so since more than 20\% (the colormap indicates the proportions) has a hamming-distance of 0, both on the X- and Y-axis. 



\section{Discussion and future work}

    Summarizing our results it seems our models perform well in approximating the level-shape input, but badly with respect to the distribution vector. Further, they perform well in the \textit{color island}-test but struggles with the \textit{broken piece}-test. Generally it must also be said that our CWGAN-GP outperforms the WGAN-GP-PE and going forward it is likely more rewardable to focus our attention towards the CWGAN-GP architecture. Let us look at the tests for condition-vectors and color-islands/broken pieces separately: \enlargethispage{13pt}
    
        \subsection{The condition vectors}
            In our generator we concatenate the \textbf{z}-vector and the \textbf{piece-\break distribution} vector. We then lead the concatenated result through some densely connected layers, before reshaping it into a (9, 15, 7) matrix, representative of the blocker- and color-layers in a level. Finally we concatenate the level-shape and the (9, 15, 7) matrix together. While we're not certain that the concatenation of the \textbf{z}-vector and \textbf{piece-distribution}-vector is the problem, it certainly hasn't proved to be a solution either. Conversely our model is adapting quite well to the level-shape, but this is also an input that can be fed in and remain non-altered before we apply convolutions, because it already has the (9, 15, 1)-shape. By contrast our \textbf{z}- and \textbf{piece-distribution}-vector has to be led through some densely connected layers in order to scale them up and into the (9, 15, 7) matrix. This could incentivize a look into other forms of representation of conditions with the closest example being Giacomello et al. (2018) which extracts a feature topology from DOOM levels, and in their case the model responded positively to this addition \cite{Giacomello2018}.  

            A possible improvement to the distribution-condition-vector might be to use some arithmetic operations on each of the layers, thus more directly connecting each cell in the condition-vector to the specific layer that it is meant to affect. This operation could be done once or on numerous occasions throughout the convolutions, so as to impose the importance of the distribution-condition-vector more explicitly.  

            A completely other avenue of inquiry, that we would also like to focus on in the future, is to focus on feedback from the intended users to understand whether our current idea of parameters should change in favour of other parameters. This would greatly improve the likelihood of integrating GANs into mixed-initiative tools that are actually likely to work for the intended users.
            
        \subsection{Color islands and broken pieces:}
            Testing for color-islands and broken pieces also give two very differing results. But it is worth considering the nature of these tests: Testing color-islands is essentially a question of finding just 1 color-layer in which 2 pieces are located next to each other. Likewise for broken pieces just 1 piece has to be out of place, but where the color-islands counts as a passing grade, the broken pieces count as a failing. In some sense they can be seen as occupying two extremes of a spectrum in which we could create others tests and it would be instructive to try and formulate some tests that better breach this divide. 
            With our current representation we are also missing an aspect of PCG regarding \textit{broken-pieces} which goes beyond what we are currently testing. In  "Evolving Mario Levels in the Latent Space of a Deep Convolutional Generative Adversarial Network" by Volz et al. they present the prevalence of broken tubes in their generated content for Mario, but in our content this simply isn't possible with our current representation but we presume that we will encounter the same issue once we introduce 2x2 pieces (such as the bushes in fig. \ref{fig:teaser}) and this consequently will also be something that needs attention in the future \cite{Volz2018}. 



\section{Conclusion}

    In this paper, we presented two GAN-architectures (WGAN-GP-PE and CWGAN-GP) for producing simplified levels of the puzzle-game Lily's Garden. To test the synthetic levels generated by the GANs, we test its ability to create levels, by testing whether the GANs are reproducing the desired level-shape we feed them correctly, as well as the piece-distribution. Both GANs reproduce the level-shape well, although the CWGAN-GP seems to be performing best, while they both fail to pick up our intended meaning for the piece-distribution-vector. We further tested the \textit{functionality} of the produced levels, by testing whether the models are producing levels with at least one \textit{color-island} in them and further whether the GANs produces any \textit{broken}-pieces. Both models perform well when tested for color islands, although the CWGAN-GP once again seems to be performing best. Conversely both models struggles when tested for broken pieces, although once again the CWGAN-GP seems to be performing better than the WGAN-GP-PE. In the future, it is therefore plausible that it is more feasible to continue with the CWGAN-GP model. However, before too much time is spend on optimizing for the variables we chose in this paper, it is worth investigating whether the condition-vectors we chose are also what game-designers might actually want or if entirely other variables might be of greater importance.


\bibliographystyle{ACM-Reference-Format}

\end{document}